\documentclass[11pt]{article}

\usepackage[final]{acl}
\usepackage{array} 

\usepackage{times}
\usepackage{latexsym}
\usepackage{comment}

\usepackage[T1]{fontenc}

\usepackage[utf8]{inputenc}

\usepackage[T1]{fontenc}

\usepackage{microtype}

\usepackage{inconsolata}

\usepackage{graphicx}
\usepackage{booktabs}
\usepackage{xcolor} 
\usepackage{multirow}
\usepackage{hhline}
\usepackage{booktabs}
\usepackage{xspace}
\usepackage{makecell}
\usepackage{vcell}
\usepackage{hyperref}
\usepackage{float}

\usepackage{pgfplots}
\pgfplotsset{compat=1.18}
\title{ALBA: A European Portuguese Benchmark for \\Evaluating Language and Linguistic Dimensions in Generative LLMs}

\author{
 \textbf{Inês Vieira\textsuperscript{1}},
 \textbf{Inês Calvo\textsuperscript{1}},
 \textbf{Iago Paulo\textsuperscript{1,2}},
 \textbf{James Furtado\textsuperscript{1,2}},
 \textbf{Rafael Ferreira\textsuperscript{1,2}},
\\
 \textbf{Diogo Tavares\textsuperscript{1,2}},
 \textbf{Diogo Glória-Silva\textsuperscript{1,2}},
 \textbf{David Semedo\textsuperscript{1,2}},
 \textbf{João Magalhães\textsuperscript{1,2}}, \\
 \textsuperscript{1}NOVA University of Lisbon, Portugal, \textsuperscript{2}NOVA LINCS \\
\texttt{\{im.vieira, i.calvo, df.semedo, jmag\}@fct.unl.pt} \\
\texttt{\{im.paulo, jh.furtado, rah.ferreira, dc.tavares, dmgc.silva\}@campus.fct.unl.pt}
}

\begin{document}

\maketitle

\begin{abstract}
As Large Language Models (LLMs) expand across multilingual domains, evaluating their performance in under-represented languages becomes increasingly important. European Portuguese (pt-PT) is particularly affected, as existing training data and benchmarks are mainly in Brazilian Portuguese (pt-BR). To address this, we introduce ALBA, a linguistically grounded benchmark designed from the ground up to assess LLM proficiency in linguistic-related tasks in pt-PT across eight linguistic dimensions, including Language Variety, Culture-bound Semantics, Discourse Analysis, Word Plays, Syntax, Morphology, Lexicology, and Phonetics and Phonology. ALBA is manually constructed by language experts and paired with an LLM-as-a-judge framework for scalable evaluation of pt-PT generated language. Experiments on a diverse set of models reveal performance variability across linguistic dimensions, highlighting the need for comprehensive, variety-sensitive benchmarks that support further development of tools in pt-PT\footnote{\url{https://github.com/AMALIA-LLM/alba-benchmark}}.
\end{abstract}

\section{Introduction}

\begin{figure} [hbt!]
    \centering
    \includegraphics[width=0.95\linewidth]{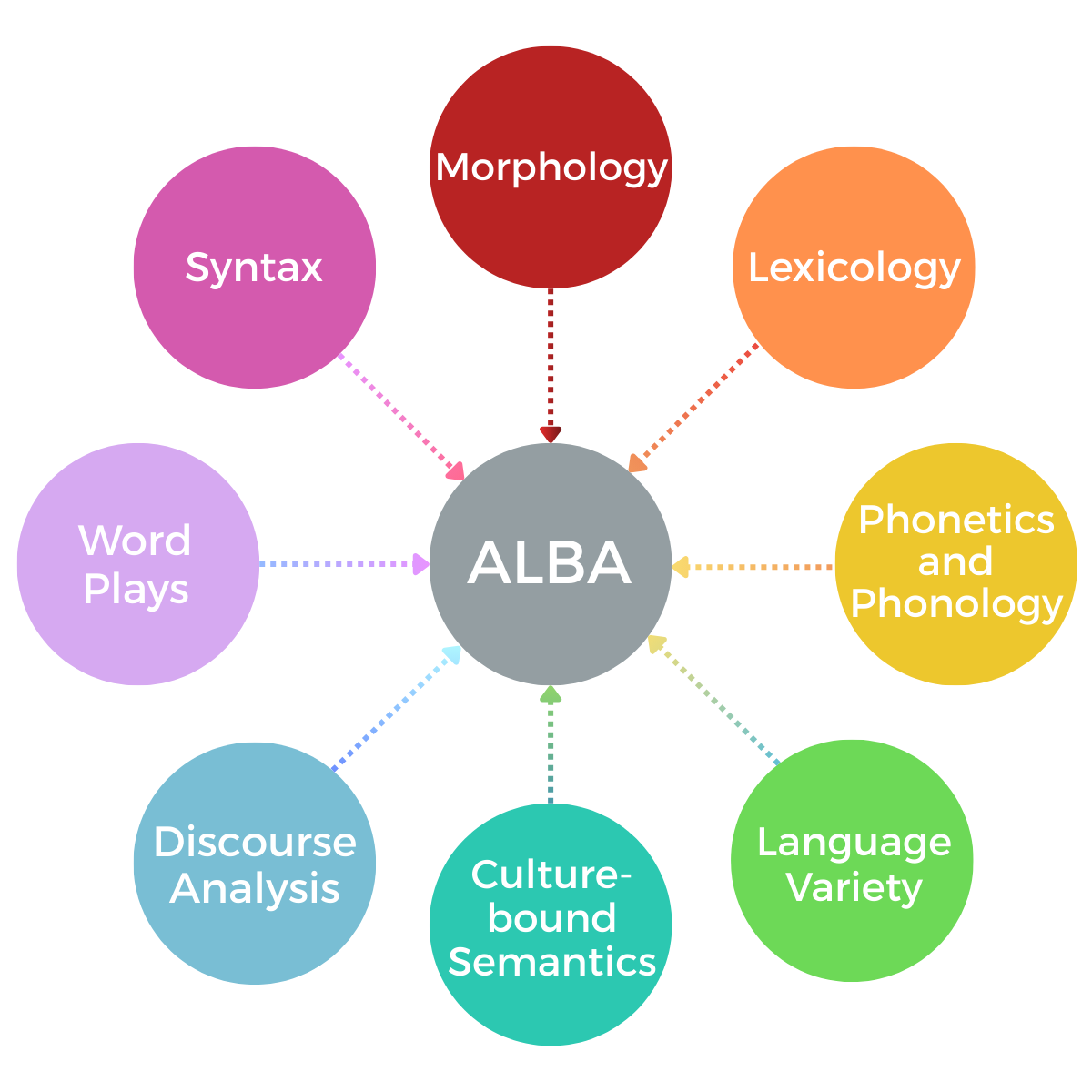}
    \caption{ALBA allows the assessment of LLM generative capabilities across eight linguistic dimensions.}
    \label{fig:alba pipeline}  
\end{figure}

While Large Language Models (LLMs) have generally progressed remarkably, their progress in lower-resource languages has been less marked~\cite{donttrustChatGPT,clickbenchmarkdatasetcultural}. 
High-resource languages form the prime focus, frequently relegating low-resource languages benchmarks to machine translation (MT) datasets.
European Portuguese (pt-PT) exemplifies this issue, as data is overwhelmingly dominated by Brazilian Portuguese (pt-BR), leading to systematic biases in which pt-PT is frequently conflated with pt-BR during both training and evaluation. As a result, various assessments provide only a partial, and often misleading, picture of LLM capabilities for the pt-PT variety.

Existing evaluation frameworks for pt-PT suffer from two major limitations. First, most benchmarks are English-centric or rely on machine translation from English~\cite{okapi_multilingual,deepl_translated_datasets}. While MT offers a scalable and convenient solution, it introduces a systemic bias that obscures language-specific phenomena such as wordplay, rhyme, or idiomatic expressions, making it unsuitable for fine-grained linguistic evaluation. 
Second, due to the under-representation of pt-PT data, most models frequently default to pt-BR lexical, morphological, or syntactic patterns, producing outputs that lack pt-PT linguistic authenticity~\cite{hate_speech_pt_pt,gloria,frmt,iberbench}.

Facing the need to assess LLMs' generative quality in pt-PT, we introduce ALBA (\textbf{A}utomated \textbf{L}inguistics \textbf{B}enchmark for baseline \textbf{A}ssessment), a benchmark manually created by domain experts that departs from standard binary/multiple-choice evaluation in favor of text generation in order to engage with language on multiple levels. In this work, inspired by the efforts done in other low-resource languages facing similar challenges~\cite{clickbenchmarkdatasetcultural, DanishNLU}, we propose a selection of eight linguistic dimensions to evaluate the linguistic quality of LLMs for European Portuguese (pt-PT): Language Variety, Culture-bound Semantics, Discourse Analysis, Word Plays, Syntax, Morphology, Lexicology, and Phonetics and Phonology (Figure~\ref{fig:alba pipeline}).

To support scalable and reliable evaluation, we further propose a rigorously validated LLM-as-a-judge framework~\cite{llm_as_judge_survey} for scoring open-ended responses. This judge was calibrated against expert annotations, ensuring alignment with native-speaker intuition while enabling systematic comparison across models.
Our contributions are threefold:
\begin{itemize}
    \item ALBA, a novel benchmark, created by language experts, specifically designed to evaluate the pt-PT linguistic capabilities of generative language models (Section~\ref{sec_dataset});
    \item An LLM-as-a-judge framework that leverages ALBA to assess the pt-PT generation quality of LLMs (Section~\ref{sec_evaluation_methodology});
    \item An extensive evaluation study of LLMs, revealing fine-grained strengths and weaknesses across ALBA's eight linguistic dimensions (Section~\ref{sec_experiments}).
\end{itemize}

By moving beyond machine translation-based datasets, ALBA offers a linguistically faithful and variety-aware framework that advances the assessment of LLM proficiency in linguistic-related tasks in pt-PT.

\section{Related Work}

\paragraph{Language Evaluation Benchmarks.}

Although benchmarks for evaluating the quality of LLM language output have been previously developed, research efforts predominantly focus on high-resource languages, while low-resource languages are often confined to machine translation benchmarks. This limitation is not unique to pt-PT, as many languages similarly lack evaluation benchmarks developed for native-language assessment.

In the case of Korean, \citet{clickbenchmarkdatasetcultural} created CLIck, a benchmark that harnesses QA pairs from examinations and textbooks to address the lack of non-translated resources and cultural nuance in existing benchmarks.
As for Danish, the Danish NLU Benchmark~\cite{DanishNLU}, created with the purpose of diagnosing and potentially remedying language and cultural biases that LLMs have in low-resource languages, has tasks involving synonymy, semantic similarity, word sense disambiguation, sentiment of words in context, entailment, and idiom interpretation.

\paragraph{Linguistics Benchmarks.}
Linguistically ground-ed evaluation has an important role in the assessment of language performance. Various benchmarks evaluate LLMs through linguistically motivated tasks across multiple languages, offering structured insights into specific aspects of model behavior.
For example, IOLBENCH~\cite{iolbench}, derived from the International Linguistics Olympiad, is primarily focused on linguistics-oriented reasoning.

Others concentrate on a single linguistic subfield or on narrowly defined task types. This is the case for PhonologyBench~\cite{PhonologyBench}, which evaluates phonological awareness in LLMs, and TACOMORE~\cite{tacomore}, a prompting framework tailored for corpus-based discourse analysis with tasks centered on keyword, collocate, and concordance analysis. In addition, other Portuguese linguistics benchmarks are unavailable in pt-PT, such as BRoverbs~\cite{BRoverbs}, which evaluates the comprehension of pt-BR proverbs.

Other works have explored the evaluation of specific linguistic competencies, including morphological generalization through compositionality in Turkish and Finnish~\cite{morphologicalgen}, as well as wordplay detection for authorship attribution in French~\cite{wordplaysinauthorshipattribution}. These approaches highlight the diversity and depth of linguistically informed evaluation, while also underscoring the language-specific nature of many linguistic phenomena.

\paragraph{European Portuguese Benchmarks.}

Evaluation benchmarks for pt-PT have been developed using a variety of approaches. Some benchmarks are derived via MT from English resources; for example, PORTULAN ExtraGLUE~\cite{PORTULAN} builds on the GLUE~\cite{glue} and SuperGLUE~\cite{SUPERGLUE}, enabling Portuguese models to be evaluated on tasks originally designed for English. Other benchmarks are manually translated to preserve subtle linguistic distinctions. BATS-PT~\cite{BATS-PT} is a manual translation of the lexicographic portion of the Bigger Analogy Test Set (BATS)~\cite{bats_dataset}, supporting analogical reasoning while retaining language-specific nuances. In parallel, several resources are conceived directly in Portuguese, either by adapting native texts or focusing on specific tasks. CALAME-PT~\cite{gloria} evaluates text completion, and ASSIN 2~\cite{real2020assin} provides manually annotated sentence pairs for semantic similarity and textual entailment. 

While translation enables rapid expansion of benchmarks across languages, many linguistic phenomena do not transfer cleanly, potentially compromising evaluation validity. This limitation has motivated the creation of resources designed natively in Portuguese. ALBA addresses this challenge by covering multiple linguistic dimensions and being developed entirely in pt-PT from the outset.

\section{ALBA Benchmark Dataset}
\label{sec_dataset}

When designing a benchmark for evaluating pt-PT, it is essential to capture the subtle nuances of the language. This creates an opportunity to tailor the benchmark specifically to pt-PT, considering both linguistic and cultural dimensions. ALBA was developed with the goal of covering a wide range of linguistic aspects, organizing questions into different branches of linguistics to grasp the finer points of language from a macroscopic perspective, so as to obtain a baseline assessment of language and linguistic performance in generative LLMs.

In specific, ALBA departs from standard binary/multiple-choice evaluation and emphasizes text generation to engage with language on multiple levels. It evaluates not only understanding and grammar, but also the ability to construct and deconstruct language creatively, as in poetry and word plays. Additionally, it incorporates culturally embedded knowledge, such as proverbs, tongue twisters, and riddles, and accounts for language variations across regions. 

The methodology used to create the questions and reference answers, as well as the reasoning behind each linguistic dimension, is detailed below.

\subsection{Methodology}
\label{sec_methodology}
Given the breadth of linguistics, priority was given to domains most relevant to LLM evaluation in pt-PT. In particular, ALBA is structured around eight core linguistic dimensions, carefully selected by language experts to capture essential aspects of the language. These dimensions encompass a broad range of linguistic capabilities: Language Variety, Culture-bound Semantics, Discourse Analysis, Word Plays, Syntax, Morphology, Lexicology, and Phonetics and Phonology.

\begin{table}[t]
\centering
\small
\resizebox{\linewidth}{!}{%
\begin{tabular}{@{}p{0.20\linewidth}p{0.80\linewidth}@{}}
\toprule
\textbf{Dimension} & \textbf{Example} \\ \midrule
Language \newline Variety & \textit{Transforma esta frase que está em português do Brasil e coloca-a em português Europeu. Respeita o registo: se houver termos informais ou gíria na frase, tenta encontrar equivalentes na língua de chegada. "Valeu, cara. A gente leva ele junto."} \\ \midrule
Culture-bound \newline Semantics & \textit{Dá-me exemplos de expressões portuguesas que mencionem chuva.} \\ \midrule
Discourse \newline Analysis & \textit{Podes criar uma frase com dois pleonasmos?} \\ \midrule
Word Plays & \textit{Dá-me um exemplo de um pangrama ou pantograma em português. Inclui todas as letras e acentos do alfabeto português europeu.} \\ \midrule
Syntax & \textit{Coloca esta frase na forma passiva: "Ele vendeu a mota."} \\ \midrule
Morphology & \textit{Cria uma lista de dez palavras com oito letras que acabem em "ão".} \\ \midrule
Lexicology & \textit{Dá-me 5 antónimos para "suave".} \\ \midrule
Phonetics \& Phonology & \textit{Escreve um poema de três versos com esquema rímico aaa (em que todos os versos rimam) com uma aliteração com "e".}\\ \bottomrule
\end{tabular}%
}
\caption{Example questions of all 8 ALBA dimensions.}
\label{tab_examples}
\end{table}

\begin{table*}[t]
\centering
\scriptsize
\begin{tabular}{@{}p{0.20\textwidth} p{0.26\textwidth} p{0.26\textwidth} p{0.20\textwidth}@{}}
\toprule
\textbf{Input Question} &
  \textbf{Correct Output (5)} &
  \textbf{Moderate Output (4)} &
  \textbf{Incorrect Output (1)} \\ \midrule
Qual é o significado da expressão "lágrimas de crocodilo"? &
  A expressão "lágrimas de crocodilo" representa lágrimas que são fingidas ou exageradas e é usada para descrever um sofrimento que não é genuíno. 
Exemplo: "Antes de chegar a polícia, aquela mulher esteve aqui meia hora aos berros a dizer que o acidente era culpa nossa. Agora está ali com lágrimas de crocodilo para ver se não perde a carta de condução." &
  A expressão "lágrimas de crocodilo" representa lágrimas que são fingidas ou exageradas e é usada para descrever um sofrimento que não é genuíno. 
Exemplo: "Antes da polícia chegar, aquela mulher esteve aqui meia hora gritando, dizendo que o acidente era culpa nossa. Agora tá ali com lágrimas de crocodilo, tentando não perder a carteira de motorista." &
A expressão "lágrimas de crocodilo" refere-se a lágrimas provocadas por medo de crocodilos. \\ \bottomrule
\end{tabular}%
\caption{Input-Output pair example from Culture-bound Semantics with accompanying expert created answers at various correctness levels.}
\label{tab_input_output_examples}
\end{table*}

\paragraph{Expert-based Reference Questions.} For each dimension, diverse and curated questions were created, resulting in a total of 800 questions, with 100 per dimension, created in two rounds. In the initial round, 30 questions per dimension were created along with corresponding answers for LLM-as-a-Judge validation. 
In the second round, 70 further questions were created.
All questions were created by two language experts (co-authors) holding Master's degrees in linguistics-related fields and with expertise in European Portuguese. Each question was authored by one expert and reviewed by the other to ensure accuracy and alignment with the intended dimension.
Example questions for each dimension are provided in Table~\ref{tab_examples}.

This approach departs from previous works that were either focused on specific linguistic reasoning capabilities \cite{DanishNLU} or took inspiration from purely test-based formats \cite{clickbenchmarkdatasetcultural} by merging the two into a dataset that aims to offer a high quality dataset that breaks complex linguistic reasoning tasks into focused components, allowing for targeted evaluation of model capabilities. 
These tasks include adjusting register or tone, proofreading, solving riddles, and composing poetry, all aligned with the eight dimensions and the pt-PT linguistic context. The following section presents each dimension in detail.

\paragraph{Expert-based Reference Answers.}
\label{sec_reference_data}
To measure LLM responses' quality and to calibrate assessment methods, the same set of language experts created a set of reference responses. 
For each of the eight linguistic dimensions, the first 30 questions from round one were used, and for each one, the experts produced three distinct responses corresponding to different quality tiers, as illustrated in Table~\ref{tab_input_output_examples}. All responses were independently rated on a 1--5 Likert scale, where 1 corresponds to an incorrect, substantially flawed and/or low language quality response, and 5 corresponds to a fully correct, complete and/or high language quality response.

This process resulted in a total of 720 expert-rated responses, which serve as ground truth for evaluating and optimizing LLM judges.

\subsection{Linguistic Dimensions}
In this section, we present ALBA's linguistic dimensions, along with a detailed explanation of what each dimension is designed to target.

\subsubsection{Language Variety}

This dimension evaluates LLMs’ ability to distinguish between pt-PT and pt-BR. It further extends this distinction to dialectal variation, defined as “a way of talking that belongs to a particular part of a country” \cite[p.~72]{littlebookoflanguage}, by targeting regionalisms.
The existence of multiple varieties of Portuguese, coupled with the resource disparity between pt-BR and pt-PT, makes it challenging to obtain an output from an LLM without the influence of pt-BR, even when using a pt-PT input.

Taking this into account, this dimension is composed of questions targeting tasks such as identifying the language variety, adapting one variety into another, recognizing the terms that flag a text or sentence as pertaining to one variety or the other, as well as identifying commonly used differing terms and expressions with the same meaning or use in the distinct varieties. Moreover, we intend to evaluate LLMs’ ability to discern the richness of pt-PT through regional variations, i.e., local words and phrases, that are specific to various parts of a country \cite[p.~72]{littlebookoflanguage}. In our case, the focus are terms and expressions used in different parts of mainland Portugal (Center, North, Trás-os-Montes, Alentejo, Algarve) and in the islands (archipelagos of Madeira and Azores).

\subsubsection{Culture-bound Semantics}

In linguistics, semantics is the study of meaning in language \cite{littlebookoflanguage}. Instead of a traditional semantical approach, ours is bound to pt-PT culture, aiming to evaluate LLMs' capacity of recognizing cultural-focused aspects, i.e., idiomatic expressions and sayings used frequently in oral communication. In addition, with ALBA, the intention is for LLMs to identify the meaning behind proverbs and idiomatic expressions in other languages and find their equivalent or parallel in pt-PT.

The aim is to evaluate the model’s ability to go beyond the literal meaning and identify the underlying message behind popular expressions, idiomatic expressions and proverbs that are intrinsic to Portuguese culture and language, as well as testing its overall knowledge of these expressions.

\subsubsection{Discourse Analysis}

"Discourse analysis is the study of what we humans do with language and how we do it." \cite{discourseanalysis}. This linguistic dimension was implemented due to the need to evaluate LLMs' capability to infer meaning and interpret longer texts. At a practical level, it refers, for example, to discursive communication and interaction, text style and text typology.

The aim is to evaluate the model’s ability to analyze, adapt and extract information from a text in tasks such as proofreading, summarization, data extraction, subject-specific text generation, text completion, and text adaptation, which are then broken down into more targeted tasks, such as the recognition, change, creation or generation of register, text typology, keywords, figures of speech, paraphrase, direct and indirect speech, divergences in content and chronological order.

\subsubsection{Word Plays}

In this dimension, language is twisted, manipulated, reconstructed and made to fit a mold through the use of word plays (such as palindromes, pangrams, isograms, marsupial words, anagrams, acrostics, calligrams, alphabet soups), with an additional focus on word and letter count, letter recognition, and rule-based sentence generation. 

Overall, the aim is to test the limits of language manipulation in tasks that usually prove too challenging for LLMs \cite{brokenwordsbrokenperformance,llmslackunderstanding}, regardless of the language being used, with the ultimate objective of evaluating how well a model can complete these tricky, convoluted tasks, if at all, without compromising language quality and fluency.

\subsubsection{Syntax}

In linguistics, syntax is the subject that studies and describes the system of rules that are followed when building sentences \cite{syntax}. From a practical standpoint, it refers to sentence categorization and correlation between words.
This ALBA dimension aims to check to which extent models are capable of understanding the language foundations, i.e., order, dependence, and hierarchy of words in sentences.

Our examples convey the following evaluation spectra: subject, verb, types of objects, other sentence constituents, types of clauses, active and passive voices, types of sentences. The overall aim of these targeted tasks is to evaluate the model’s ability to complete more overreaching complex syntax-related exercises, such as proofreading, which makes use of textual fine-tuning and overall rephrasing for clarity and textual enhancement, as well as text and sentence analysis.

\subsubsection{Morphology}

In linguistics, morphology "is the study of word formation, including the ways new words are coined in the languages of the world, and the way forms of words are varied depending on how they're used in sentences." \cite[p.~9]{morphology} and stands for the description and analysis of words’ internal structure. In practice, it refers, for example, to word inflection according to number, gender, and tense.

This dimension aims to check to which extent LLMs are capable of understanding and replicating word formation and inflection in pt-PT, so as to enable higher-complexity related tasks (e.g. changing the register of a text by changing the verb tenses from first person singular to third person singular in pt-PT, and vice-versa). Our examples convey the following evaluation spectra: verb tenses, adjective degrees, morphological constituents (i.e., affixes, interfixes, suffixes), prepositions, thematic vowels.

\subsubsection{Lexicology}

Lexicology stands for the linguistic dimension that studies "lexis, understood as the stock of words in a given language, i.e., its vocabulary or lexicon" \cite[p.~1]{lexicology}. Thus, it analyses words, including their origin and morphological, syntactical and semantic features, as well as describing processes at the basis of new words’ formation. This dimension aims to check to which extent LLMs are capable of understanding relations between words, particularly their lexical mastery range, hierarchy and lexical network. 

In order to better perceive how a model might perform in tasks, such as error correction, proofreading, changes in clarity and conciseness, as well as lexical richness, these tasks are dissected into their most fundamental parts so as to extract the knowledge and skills which they rely on. Therefore, this dimension focuses on the following evaluation spectra: synonymy, antonymy, homophony, homography, hyponymy, hypernymy, lexical field, semantic field, word family, neologism, archaism.

\subsubsection{Phonetics and Phonology}

Phonetics and Phonology were grouped in one dimension, since, from distinct perspectives, both subjects study language sounds and, as a result, they are complementary.
Phonetics is the study of speech sounds \cite{littlebookoflanguage}, meaning the subject that studies and describes the physical, perceptive, articulatory, and acoustic features of speech sounds.
On the other hand, phonology is the study of sound structure in language \cite{phonology}, meaning the subject that studies languages’ speech sounds and sound patterns.

This ALBA dimension aims to check to which extent models are capable of understanding structure and internal constraints of language, while assessing their sensitivity to abstract representation. Our dataset conveys the following evaluation spectra: vowels, consonants, diphthongs, hiatus, syllables classification, alliterations, tongue twisters, rhyme types, rhyming words, scansion, phonetic transcription, sounds repetition, and speech sounds classification.

The aim is to evaluate the model's understanding of sound as it relates to the language being evaluated and its ability to perform creativity and language related tasks, such as the creation of poems and tongue twisters, as well as literature and language analysis related tasks (e.g. scansion), while breaking down the necessary components for completing these kinds of tasks, such as rhyme, accentuation, metrics, and structure.

The idea behind this is that if a model is not able to successfully complete the Phonetics and Phonology related tasks that have been broken down into individual exercises (e.g. rhyme, metrics, structure), it will not be able to do more complex tasks (e.g. create an original sonnet or analyze the rhyme scheme of a poem).

\section{ALBA's LLM-as-a-Judge Framework}
\label{sec_evaluation_methodology}

Given the complexity of evaluating linguistic competence across multiple dimensions and the subjective nature of assessing open-ended responses, we adopted an LLM-as-a-judge approach for evaluating model outputs~\cite{llm_as_judge_survey}. 
This methodology allows for scalable evaluation while maintaining the nuanced understanding required for linguistic assessment.

Our judge selection methodology consists of three stages:
\begin{enumerate}
    \item Creation of expert-annotated reference responses (Section~\ref{sec_methodology});
    \item Systematic exploration of prompt configurations and judge models;
    \item Selection of the most reliable judge and configuration for benchmark evaluation.
\end{enumerate}

\subsection{LLM Judge Configuration}
\label{sec_judge_configuration}

We systematically explored the LLM-as-a-judge configuration space to identify the setup that best aligns automated evaluations with human judgments. In particular, we examined the effects of prompt language, few-shot example selection, and the choice of judge model.

\paragraph{Prompt Design.} The judge prompt assigns the model the role of a professional text evaluator and receives a detailed expert-defined rubric with a 1 (Very Bad) to 5 (Very good) scoring scale, covering precision, linguistic quality, and completeness. When few-shot examples are used, they are appended after the rubric to guide scoring. The judge is instructed to first produce an explicit reasoning trace using a chain-of-thought~\cite{cot_paper} before providing the final numeric score, a process shown to improve evaluation reliability and consistency~\cite{g_eval,dhp_eval}.

\begin{figure*}[t]
  \centering
  \begin{minipage}{0.48\linewidth}
    \centering
    \includegraphics[width=\linewidth]{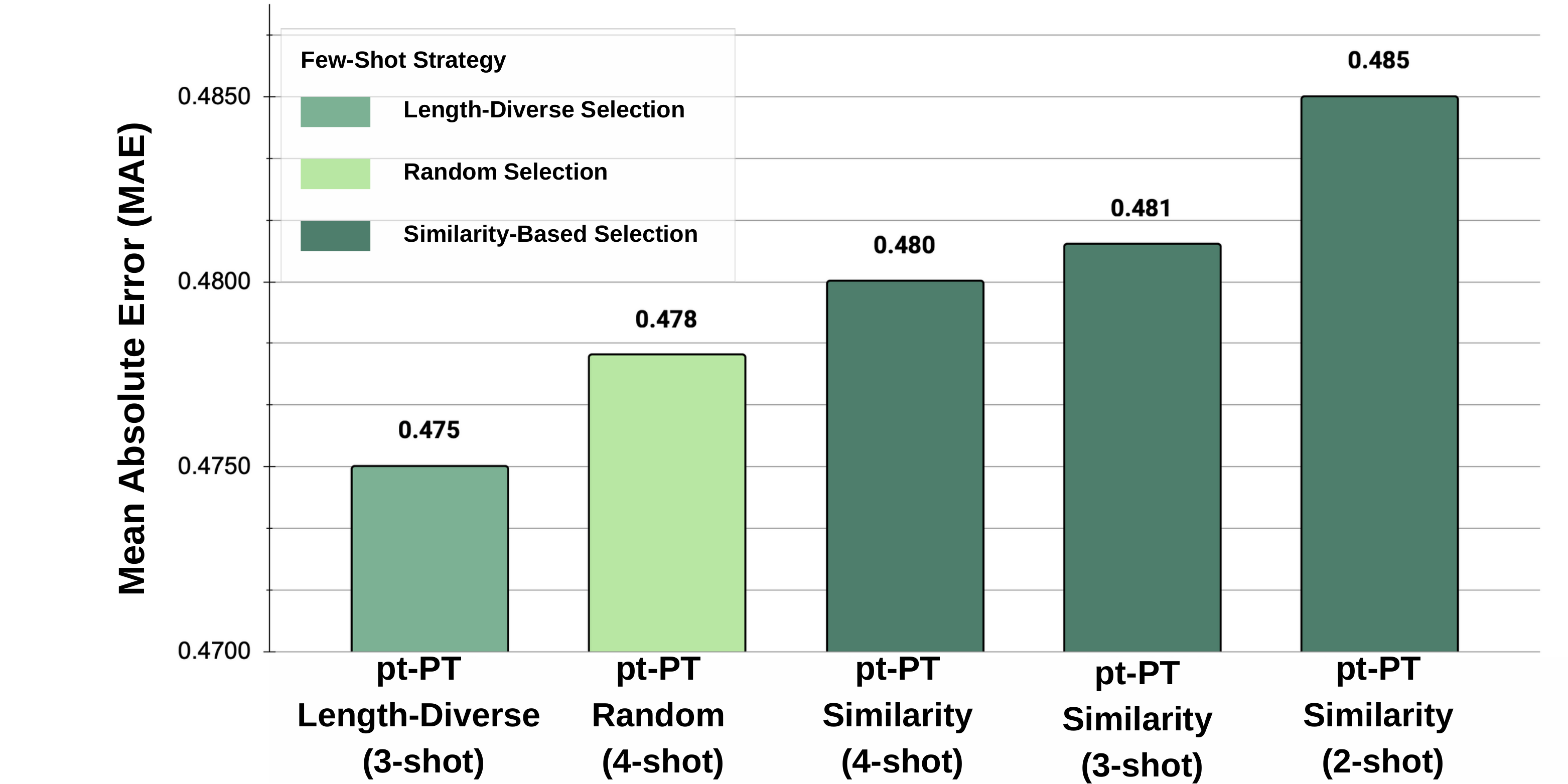}
    \caption{Top five judge configurations ranked by MAE on the validation set.}
    \label{fig:top_configs}
  \end{minipage}
  \hfill
  \begin{minipage}{0.48\linewidth}
    \centering
    \includegraphics[width=\linewidth]{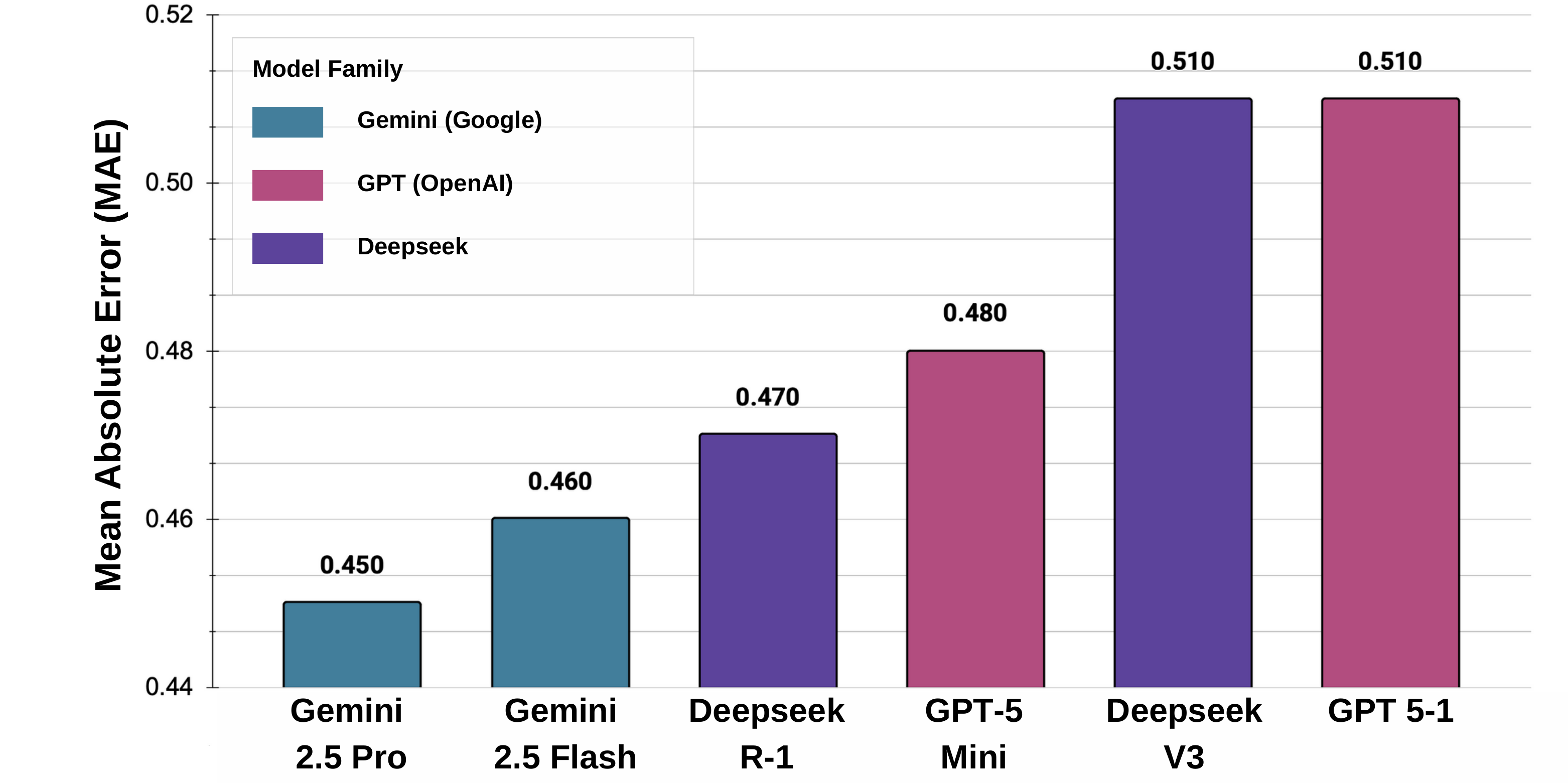}
    \caption{MAE of different LLM judges on the evaluation set using the optimal configuration.}
    \label{fig:model_comparison}
  \end{minipage}
\end{figure*}

\paragraph{Prompt Language.}
We tested prompts written in pt-PT and English (EN) to evaluate whether native-language prompting improves judgment accuracy for pt-PT content.

\paragraph{Few-Shot Examples.}
We evaluate the effect of few-shot prompting on evaluation quality by varying both the number of examples (2--5 samples per prompt) and the example selection strategy~\cite{few_shot_learning}. Each example comprises three candidate responses reflecting distinct quality levels: correct, moderate, and incorrect (Section~\ref{sec_methodology}).  
For example selection, we consider the following strategies:
\begin{itemize}
    \item \textbf{Random}: samples chosen at random;
    \item \textbf{Similarity}: selecting semantically dissimilar samples to maximize diversity;
    \item \textbf{Length-Diverse}: samples with maximally different response lengths (e.g., shortest, median, longest) are selected to increase diversity in structure.
\end{itemize}

\paragraph{Candidate Judge Models.}
In addition to varying the prompt, we evaluated multiple LLMs as candidate judges. The selected models, Gemini-2.5~\cite{gemini_2_5}, DeepSeek~\cite{deepseek_r1,deepseek_v3}, and GPT-5~\cite{openai2025gpt5card}, were chosen for their strong reasoning capabilities, robustness in zero- and few-shot settings, and multilingual understanding. 

\subsection{Experimental Protocol}
From the expert-rated inputs (Section~\ref{sec_reference_data}), we created two disjoint subsets. In particular, we used two-thirds of the data as a validation set to optimize the judge prompt (language, and few-shot configuration), while the remaining served as a held-out evaluation set for selecting the LLM judge model.

Judge performance was measured by the Mean Absolute Error (MAE) between the scores given by the LLM judge and the expert ratings, where a lower MAE indicates closer alignment with human judgment.

\subsection{Results}
\label{sec_optimal_configuration}

\paragraph{Prompt and Few-Shot Configuration.}
Figure~\ref{fig:top_configs} presents the five best-performing configurations on the validation set. Across all cases, the results consistently favor the pt-PT prompt.
In particular, the configuration using three few-shot samples selected with the length-diverse strategy achieved the lowest MAE ($0.475$), demonstrating a good alignment with expert ratings.

\paragraph{Model Selection.}
Using the optimal configuration identified in the previous section, Figure~\ref{fig:model_comparison} reports the results for each candidate model. The results show that Gemini-2.5-Pro achieves the lowest MAE, emerging as the most reliable judge.

Based on these results, the final LLM judge setup employs Gemini-2.5-Pro with a pt-PT prompt and three few-shot samples.
Generation is performed using greedy decoding..

\begin{table*}[t]
\centering
\resizebox{\textwidth}{!}{%
\begin{tabular}{@{}lc|cccccccc@{}}
\toprule
\textbf{Model} &
  \textbf{ALBA} &
  \textbf{\begin{tabular}[c]{@{}c@{}}Language\\ Variety\end{tabular}} &
  \textbf{\begin{tabular}[c]{@{}c@{}}Culture-bound\\ Semantics\end{tabular}} &
  \textbf{\begin{tabular}[c]{@{}c@{}}Discourse\\ Analysis\end{tabular}} &
  \textbf{\begin{tabular}[c]{@{}c@{}}Word\\ Plays\end{tabular}} &
  \textbf{Syntax} &
  \textbf{Morphology} &
  \textbf{Lexicology} &
  \textbf{\begin{tabular}[c]{@{}c@{}}Phonetics \&\\ Phonology\end{tabular}} \\ \midrule
\textit{Fully open models}  & & & & & & & & & \\
OLMo 2-7B~\cite{olmo2} &
  16.9 &
  9.5 &
  9.5 &
  43.3 &
  4.0 &
  24.5 &
  26.0 &
  14.0 &
  4.8 \\
Salamandra-7B~\cite{gonzalez2025salamandra} &
  27.4 &
  27.8 &
  30.5 &
  52.8 &
  4.8 &
  26.5 &
  36.3 &
  29.3 &
  11.5 \\
EuroLLM-9B~\cite{eurollm} &
  38.5 &
  41.0 &
  40.0 &
  67.0 &
  11.3 &
  43.3 &
  44.0 &
  47.5 &
  13.8 \\
Apertus-8B~\cite{apertus} &
  38.7 &
  37.8 &
  38.3 &
  70.3 &
  13.8 &
  47.5 &
  45.0 &
  45.8 &
  11.5 \\
AMALIA-9B~\cite{amalia} &
  43.6 &
  48.3 &
  \underline{47.8} &
  73.0 &
  14.8 &
  42.3 &
  49.8 &
  \underline{53.8} &
  19.0 \\ \midrule
\textit{Open weight models}  & & & & & & & & & \\
Mistral-7B~\cite{mistral7b} &
  21.7 &
  15.5 &
  17.5 &
  50.0 &
  3.5 &
  32.5 &
  26.3 &
  26.8 &
  1.8 \\
Ministral-8B~\cite{ministral8b2024} &
  35.6 &
  32.0 &
  37.5 &
  61.0 &
  12.0 &
  45.8 &
  50.8 &
  34.0 &
  11.8 \\
LLaMA 3.1-8B~\cite{grattafiori2024llama} &
  31.3 &
  27.8 &
  23.5 &
  60.5 &
  19.3 &
  35.0 &
  38.5 &
  29.3 &
  17.0 \\
Gervasio-8B~\cite{gervasio} &
  31.1 &
  29.0 &
  22.8 &
  61.8 &
  17.0 &
  38.3 &
  39.5 &
  25.8 &
  15.0 \\
Qwen 2.5-7B~\cite{qwen2.5} &
  31.0 &
  26.3 &
  24.0 &
  63.0 &
  11.0 &
  42.8 &
  38.3 &
  32.3 &
  10.3 \\
Qwen 3-8B~\cite{qwen3technicalreport} &
  49.8 &
  44.8 &
  35.8 &
  77.5 &
  31.0 &
  \underline{70.3} &
  53.8 &
  44.5 &
  \underline{41.0} \\
Gemma 2-9B~\cite{gemma2} &
  41.1 &
  40.8 &
  35.5 &
  78.5 &
  22.8 &
  44.8 &
  48.5 &
  41.3 &
  16.8 \\
Gemma 3-12B~\cite{gemma_2025} &
  \underline{51.1} &
  \underline{52.8} &
  41.5 &
  \underline{85.5} &
  \underline{34.3} &
  58.0 &
  \underline{56.8} &
  50.8 &
  29.5 \\ \midrule
\textit{Close source models} & & & & & & & & & \\
GPT-5~\cite{openai2025gpt5card} &
  \textbf{91.0} &
  \textbf{98.3} &
  \textbf{88.3} &
  \textbf{98.5} &
  \textbf{91.5} &
  78.3 &
  \textbf{98.5} &
  88.8 &
  \textbf{86.3} \\
Gemini 2.5-Pro~\cite{gemini_2_5} &
  90.1 &
  96.5 &
  85.3 &
  98.0 &
  85.3 &
  \textbf{85.8} &
  95.8 &
  \textbf{89.5} &
  84.8
  \\ \bottomrule
\end{tabular}%
}
\caption{Performance of models on ALBA across linguistic dimensions, based on LLM-as-a-judge evaluation with original 1--5 ratings rescaled to a 0--100 range. Bold indicates the best overall model, while underlining denotes the best-performing non-closed-source model.}
\label{tab_alba_gen_results}
\end{table*}

\section{Experiments}
\label{sec_experiments}
In this section, we analyze the performance of multiple LLM models in terms of language and linguistic generation capabilities according to ALBA's dimensions and the LLM-as-a-judge framework.

\subsection{Baseline Language Models}
\label{sub_models}

We evaluate multilingual instruction-tuned LLMs with 7B--12B parameters, including fully open-source and open-weight models from major families (e.g., OLMo, Mistral, Gemma, Qwen, and LLaMA). Models were selected based on strong reported Portuguese and multilingual performance, public availability, and architectural diversity, while controlling for model scale to enable fair comparison.
We additionally include GPT-5 and Gemini-2.5-Pro\footnote{Using the same model as the judge may introduce potential bias~\cite{llm_judge_same_bias}, however, our judge validation showed good alignment with human judgments.} as frontier baselines, contextualizing open-weight results against current proprietary state-of-the-art systems.

\subsection{Overall Results}
Table~\ref{tab_alba_gen_results} presents the performance of instruction-tuned models on the ALBA benchmark.

Overall, fully open models achieve lower scores. OLMo-2 performs poorly due to its monolingual (English-only) training. Multilingual models such as Euro-LLM and Apertus-8B perform better, but still lag behind the strongest open-weight models.
AMALIA, tailored for European Portuguese (pt-PT), achieves the strongest results among fully open models, even outperforming the larger Gemma 3-12B in culturally bound semantics and lexicology, highlighting the benefits of language-specific specialization.
The Gemma models show good results, with Gemma 3-12B achieving the highest score ($51.1$) among open-weight models, showing a good understanding of pt-PT linguistics. Qwen 3-8B follows closely ($49.8$), delivering competitive performance despite its smaller size, likely due to its explicit reasoning capabilities. 
However, all open models exhibit clear limitations in more fine-grained dimensions. In Phonetics and Phonology, errors frequently involve metric inconsistencies, rhyme misclassifications, tonic stress misplacement, and phonetic hallucinations. In Word Plays, models often produce word hallucinations or fail to correctly manipulate characters.

\begin{figure*}[t]
    \centering
    \includegraphics[width=1\linewidth]{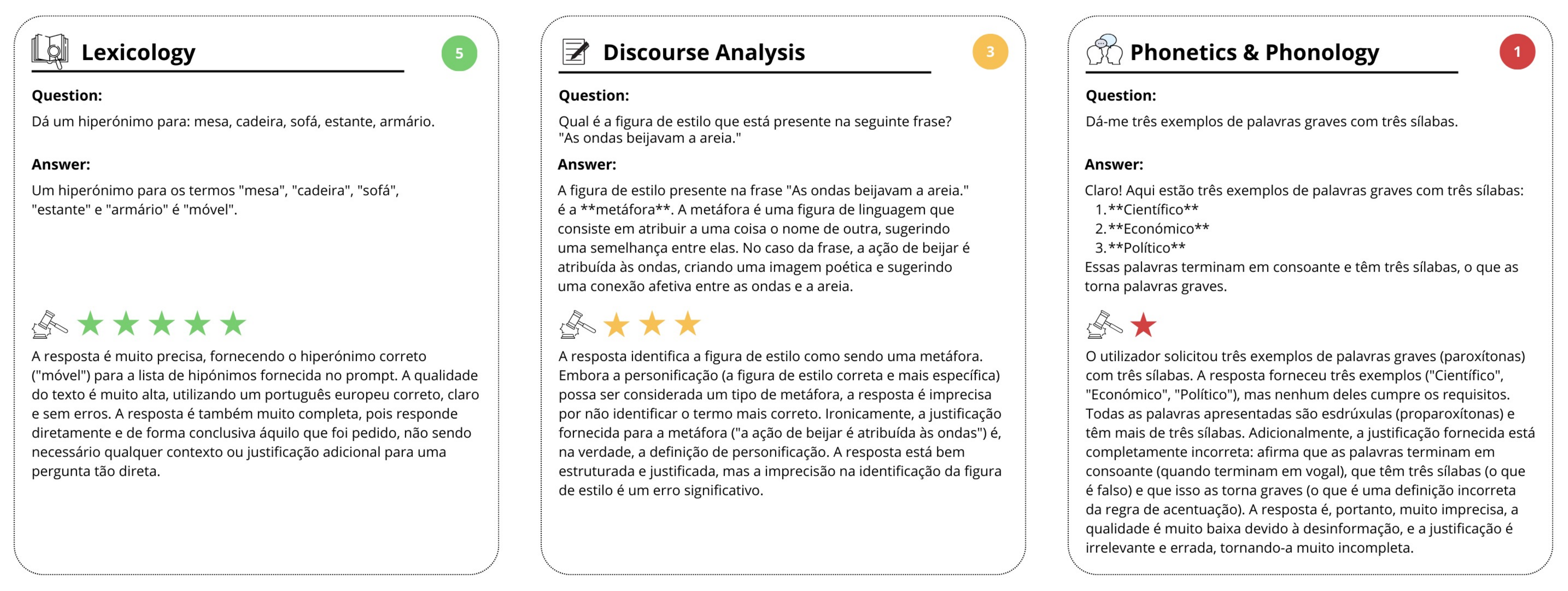}
    \caption{Illustrative answers and judge evaluations from Ministral on various ALBA dimensions.}
    \label{fig:comparison}
\end{figure*}

Even in the dimensions where open models performed better, there were still recurring errors. In Culture-bound Semantics, models often failed to recognize or generate culturally intrinsic elements and humor. In Language Variety, they hallucinated slang or regional terms and confused language varieties. In Discourse Analysis, common issues included misidentification of rhetorical devices and difficulty detecting irony. Terminological inconsistencies and unintended language mixing were also frequent across dimensions.
In Figure \ref{fig:comparison}, we show the model and judge outputs for three different examples.

From a scaling perspective, a clear gap remains between open-weight and closed-source models. Closed-source models reach consistently high performance across all dimensions, including the more complex ones, benefiting from larger scale and broader training data.

\subsection{Results Analysis}

The results obtained on ALBA are consistent with previously established trends in the evaluation of LLMs on linguistic tasks. As observed in other benchmarks \cite{tacomore, DanishNLU, holmesbenchmarkassesslinguistic}, LLMs tend to perform well on syntactic, lexical, and discourse-level tasks, while exhibiting substantially lower performance in other linguistic dimensions.

As previously established \cite{PhonologyBench, morphologicalgen, llmslackunderstanding}, LLMs exhibit limitations in tasks involving phonology, morphology, and wordplay, particularly in tasks involving rhyme, scansion, syllable segmentation, morphological composition, and character-level manipulations such as reordering or counting.

This contrast in performance may be attributed to the way LLMs process natural language, specifically the segmentation of words into tokens. Although tokenization can enhance performance on tasks involving syntax and grammatical structures \cite{unveilingmakeslinguisticsolympiad, holmesbenchmarkassesslinguistic, blimpbenchmarklinguisticminimal}, it can adversely affect others, such as word insertion and retrieval, as well as character counting, insertion, and deletion \cite{llmslackunderstanding}.

The results in ALBA further substantiate previously reported discrepancies in the linguistic performance of LLMs, extending prior findings from linguistic benchmarks to the context of pt-PT.

\section{Conclusion}

In this work, we introduced ALBA, a linguistically grounded benchmark for pt-PT that encompasses structural, semantic, cultural, and variety-sensitive aspects of the language. Unlike previous benchmarks, ALBA broadens the scope of tasks to provide a more comprehensive evaluation of language generation and linguistic competence. ALBA includes eight linguistic dimensions and 800 expert-crafted questions, supported by a validated LLM-as-Judge framework.

Our results show that current open models perform better on straightforward tasks such as Discourse Analysis and Syntax but struggle with more intricate areas, including Phonetics \& Phonology and Word Plays, underscoring the importance of diverse, linguistically grounded data.

In summary, ALBA provides a language attuned framework to measure proficiency in linguistic-related tasks in pt-PT. Future work should expand these capabilities to additional under-represented languages and linguistic phenomena, enhancing its coverage, relevance, and utility.

\section*{Acknowledgments}
This work was supported by the AMALIA project under Measure RE-C05-i08 of the Portuguese national Programa de Recuperação e Resiliência. We also acknowledge the support of Fundação para a Ciência e Tecnologia (FCT) and the NOVA LINCS project (UID/04516/2025). Finally, we thank the Barcelona Supercomputing Center (BSC) for providing the computational resources that made this work possible.

\bibliography{references}

\end{document}